\documentclass[conference]{IEEEtran}

\IEEEoverridecommandlockouts

\usepackage{cite}
\usepackage{amsmath,amssymb,amsfonts}
\usepackage{algorithmic}
\usepackage{graphicx}
\usepackage{textcomp}
\usepackage{subcaption}
\usepackage{multirow}
\usepackage{booktabs}
\usepackage{xcolor}
\usepackage{float}
\usepackage{stfloats}
\def\BibTeX{{\rm B\kern-.05em{\sc i\kern-.025em b}\kern-.08em
    T\kern-.1667em\lower.7ex\hbox{E}\kern-.125emX}}

\newcounter{appendixsection}
\renewcommand{\theappendixsection}{\Alph{appendixsection}.}

\newcommand{\appendixsection}[1]{%
  \refstepcounter{appendixsection}%
  \medskip
  \noindent\textbf{\theappendixsection\ #1}\par
  \medskip
}
\begin{document}

\title{HFBRI-MAE: Handcrafted Feature Based Rotation-Invariant Masked Autoencoder for 3D Point Cloud Analysis\\}

\author{\IEEEauthorblockN{Xuanhua Yin, Dingxin Zhang, Jianhui Yu, Weidong Cai}}
\author{Xuanhua Yin, Dingxin Zhang, Jianhui Yu, Weidong Cai\\
    School of Computer Science, The University of Sydney\\
    {\tt \{xyin0811,dzha2344,jianhui.yu,tom.cai\}@sydney.edu.au} 
}

\maketitle

\begin{abstract}
Self-supervised learning (SSL) has demonstrated remarkable success in 3D point cloud analysis, particularly through masked autoencoders (MAEs). 
However, existing MAE-based methods lack rotation invariance, leading to significant performance degradation when processing arbitrarily rotated point clouds in real-world scenarios. 
To address this limitation, we introduce Handcrafted Feature-Based Rotation-Invariant Masked Autoencoder (HFBRI-MAE), a novel framework that refines the MAE design with rotation-invariant handcrafted features to ensure stable feature learning across different orientations.
By leveraging both rotation-invariant local and global features for token embedding and position embedding, HFBRI-MAE effectively eliminates rotational dependencies while preserving rich geometric structures.
Additionally, we redefine the reconstruction target to a canonically aligned version of the input, mitigating rotational ambiguities.
Extensive experiments on ModelNet40, ScanObjectNN, and ShapeNetPart demonstrate that HFBRI-MAE consistently outperforms existing methods in object classification, segmentation, and few-shot learning, highlighting its robustness and strong generalization ability in real-world 3D applications.
\end{abstract}


\section{Introduction}

The point cloud is a widely used representation of 3D data~\cite{b75}, which plays a crucial role in fields that require real-world perception, such as autonomous driving~\cite{b46}, medical data processing~\cite{b47,b70}, and robotics~\cite{b48}.
In recent years, there have been increasing attempts to apply cutting-edge self-supervised learning (SSL) frameworks to 3D point cloud recognition~\cite{b20,b21,b22,b23,b24,b25,b26,b27,b28,b44,b45}. 
As unlabeled 3D data is vastly more plentiful compared to fully annotated data, SSL approaches exploit pre-training on unlabeled data, reducing annotation costs while simultaneously enhancing performance in downstream tasks~\cite{b20}.
This demonstrates the high potential and application value of SSL in point cloud research.

Among all SSL methods, the reconstruction-based Masked Autoencoder (MAE)~\cite{b25,b26,b27} has shown remarkable effectiveness and applicability in point cloud analysis. 
By reconstructing masked point patches using partially visible ones, MAE enables the extracted features to align well with the local geometry and global context information required for point cloud recognition. 
However, existing studies remain largely confined to laboratory conditions where objects are aligned in fixed canonical poses, neglecting the arbitrary rotations encountered in real-world scenarios.
Rotation-invariant models are inherently resistant to the effects of point cloud rotations, enabling them to capture rotation-invariant features more effectively and handle complex rotational variations with greater accuracy.
Conventional MAE models, even with extensive data augmentation, experience significant performance degradation when encountering randomly rotated point clouds, often performing much worse than supervised models specifically designed for rotation invariance~\cite{b7,b8,b9,b10,b11,b12,b16,b17,b18,b19,b30,b43,b71}. 

\begin{figure}[!t]
    \centering
    \includegraphics[width=1\columnwidth]{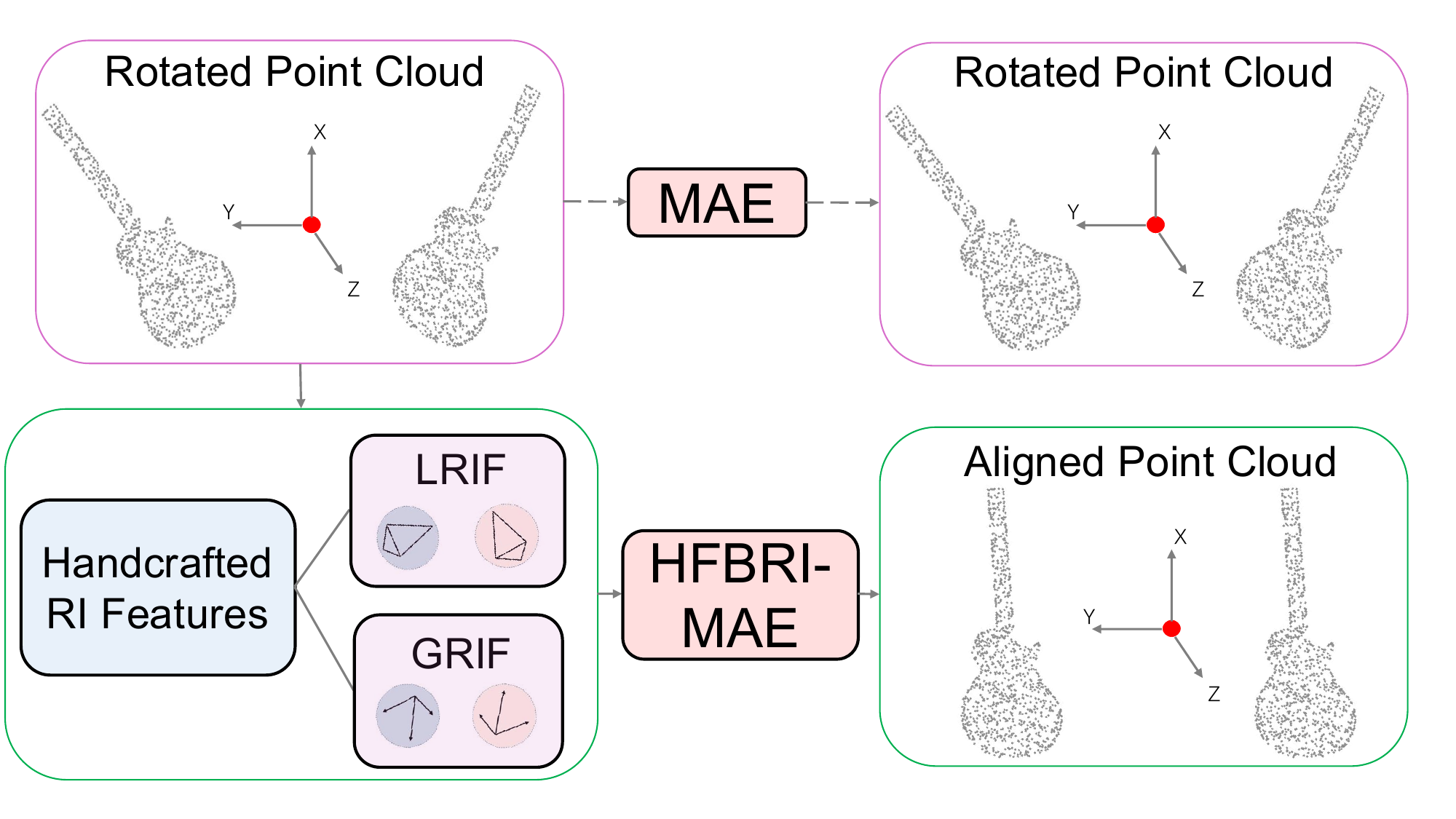}
    \caption{Comparison between standard MAE and HFBRI-MAE under rotated point clouds. HFBRI-MAE uses handcrafted rotation-invariant features and aligned reconstruction targets to achieve consistent features learning across rotations.}
    \label{fig:introduction}
\end{figure}
To address this limitation, we propose an enhancement to existing MAE models that retains the advantages of SSL while achieving rotation invariance. 
A well-established approach in self-supervised rotation-invariant point cloud models involves leveraging handcrafted features based on rotation-invariant line segments and angle information~\cite{b7,b8}.  
We hypothesize that it is possible to build a rotation-invariant MAE model by leveraging handcrafted features as input, as rotational information is eliminated before data are fed into the model.
In addition, the MAE patch masking and reconstruction mechanism forces the model to infer missing patch information from visible patches, making it well-suited for learning high-quality hidden inter-patch local and intra-patch global rotation-invariant representations for 3D point clouds.
Both token embedding and position embedding procedures can be effectively implemented using these handcrafted features.
One challenge of this approach is that MAE relies on known information to infer missing data, therefore losing orientation cues may disrupt this inference process. 
To mitigate this, we propose a carefully balanced strategy incorporating handcrafted feature design, optimal mask ratios, and well-calibrated encoder-decoder architecture to ensure the pretrained model extracts high-quality features.

Building on these insights, we introduce the \textbf{H}andcrafted \textbf{F}eature-\textbf{B}ased \textbf{R}otation-\textbf{I}nvariant \textbf{M}asked \textbf{A}uto\textbf{e}ncoder (HFBRI-MAE). 
As shown in Fig.~\ref{fig:introduction}, HFBRI-MAE eliminates rotational information from input by integrating rotation-invariant local features (RILF) and rotation-invariant global features (RIGF).
Specifically, RILF captures fine-grained geometric details for token embedding, while RIGF encodes spatial relationships among patches for position embedding.
These handcrafted features incorporate comprehensive rotation-invariant distance and angle information, enriching the feature space and enhancing the model’s ability to capture intricate geometric structures. 
Additionally, the rotational information in the output should also be removed, so we align the target reconstructed point cloud with a canonical pose, ensuring that the entire network focuses exclusively on reconstructing rotation-invariant features.
Experimental results demonstrate that HFBRI-MAE consistently outperforms existing methods across diverse tasks, including classification, segmentation, and few-shot learning. 
These findings underscore its ability to extract rotation-invariant features and highlight its adaptability for various 3D point cloud applications.
The main contributions of this work are summarized as follows:  
\begin{itemize}  
    \item We propose HFBRI-MAE, a novel point cloud SSL framework that integrates rotation-invariant handcrafted features with masked autoencoder, ensuring consistency across varying orientations.  
    \item The framework incorporates RILF and RIGF, capturing both fine-grained geometric details and holistic spatial relationships to improve representation quality.
    \item Extensive experiments on benchmark datasets demonstrate the effectiveness of HFBRI-MAE outperforms state-of-the-art methods across multiple downstream tasks, including object classification, segmentation, and few-shot learning.
\end{itemize}

\section{Related Works}

\subsection{Deep Learning in Point Cloud}

Deep learning has revolutionized 3D point cloud analysis through two dominant paradigms: MLP-based methods~\cite{b1, b2, b3} and attention-based architectures~\cite{b4, b5}. 
The pioneering work of PointNet~\cite{b1} introduced point-wise MLPs and max-pooling to aggregate global features but struggled to capture fine-grained local geometric details. 
This limitation was partially addressed by PointNet++~\cite{b2}, which employed hierarchical feature learning through point down-sampling. 
Attention-based methods, such as PCT~\cite{b4} and Point Transformer~\cite{b5}, further advanced the field by modeling dynamic point relationships. PCT leverages global self-attention for comprehensive context capture, while Point Transformer focuses on local neighborhoods to balance computational efficiency. 
In parallel, CurveNet~\cite{b73} introduced curve-based feature aggregation to enhance local geometric continuity. 
Recent advancements like PointNeXt~\cite{b3} integrate lightweight MLP designs with advanced data augmentations, achieving scalability and state-of-the-art performance.

\subsection{Rotation-Invariant Point Cloud Analysis}

Achieving rotation invariance remains a critical challenge for real-world 3D applications. 
Early efforts include handcrafted methods such as RIConv~\cite{b7}, which encodes local geometric relationships to construct rotation-invariant features, and its improved variant RIConv++~\cite{b8}, which introduces global features for enhanced robustness. 
Alternative approaches like GCANet~\cite{b9}, LGR-Net~\cite{b10} and Yu et al.~\cite{b72} rely on local reference frames (LRFs) to align point coordinates, while PaRI-Conv~\cite{b11} and PaRot~\cite{b12} disentangle pose information from geometric features. 
PCA-based methods~\cite{b16, b17} align point clouds into canonical poses but suffer from sign ambiguities and instability under noise. 
Equivariant models, such as Tensor Field Network~\cite{b18} and SE(3)-Transformer~\cite{b19}, leverage spherical harmonics and attention mechanisms to ensure feature consistency across SE(3) transformations, offering theoretically guaranteed rotation invariance.

\subsection{Self-Supervised Learning for Point Clouds}

In self-supervised learning (SSL), contrastive methods like PointContrast~\cite{b20} and CrossPoint~\cite{b21} learn representations by contrasting positive and negative pairs, though their reliance on precise alignment limits robustness. 
Clustering-based approaches~\cite{b22} group points into semantic clusters but struggle with 3D data variability. 
Reconstruction-based methods, including FoldingNet~\cite{b23} and PSG-Net~\cite{b24}, focus on structural relationships, while PointMAE~\cite{b25} and PointM2AE~\cite{b26} extend masked autoencoders (MAEs) to reconstruct masked regions. 
However, these methods lack inherent rotation invariance. 
Recent attempts like MaskLRF~\cite{b27} integrate LRFs for rotation invariance but incur prohibitive computational overhead, and RI-MAE~\cite{b28} adopts PCA-based alignment but remains sensitive to noise and point density variations.

Our work addresses these gaps by integrating rotation-invariant handcrafted features into the MAE framework. 
Unlike LRF-based methods, the proposed HFBRI-MAE avoids computational bottlenecks, and compared to PCA-based approaches, it eliminates instability caused by noisy alignments, achieving robust performance under arbitrary rotations.

\section{Methodology}
\subsection{Problem Description}
\begin{figure*}[htbp]
    \centering
    \includegraphics[width=\textwidth]{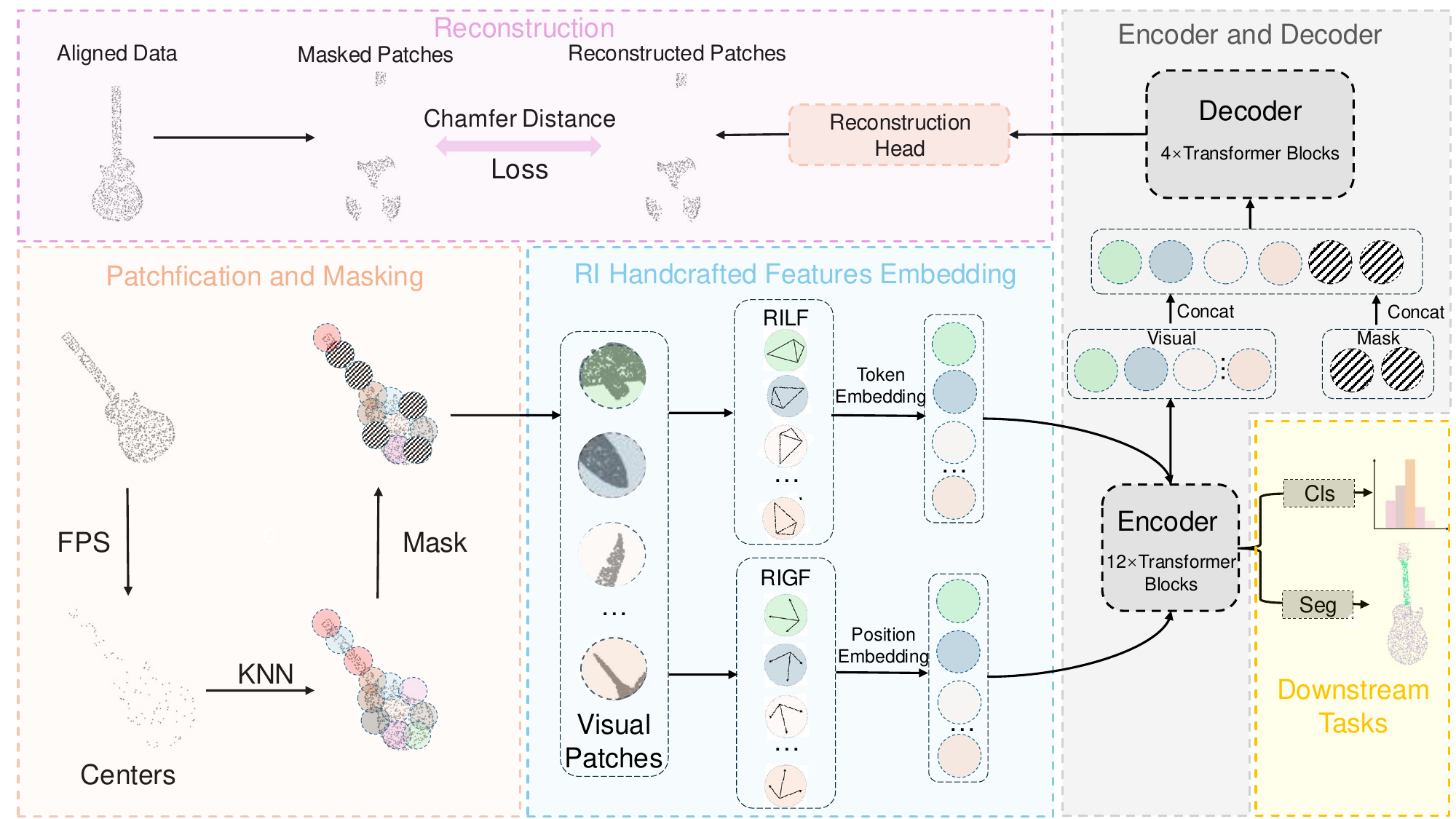}
    \caption{Architecture of the proposed HFBRI-MAE framework. The input point cloud is divided into patches using FPS and KNN. RILF and RIGF are extracted to form token and position embeddings, which are processed by the encoder. The decoder reconstructs masked patches using aligned point cloud coordinates, facilitating self-supervised learning and downstream tasks.}
    \label{fig:structure}
\end{figure*}

Rotation invariance (RI) is essential for practical 3D point cloud analysis, as real-world objects can appear in arbitrary orientations. 
Formally, a model is rotation-invariant if its feature encoder $f$ satisfies:
\begin{equation}
\label{eq:invariance}
f(\mathbf{P}) =f(\mathbf{R}\mathbf{P}), \quad \forall \mathbf{R} \in \text{SO}(3), \mathbf{P} \in \mathbb{R}^{N \times 3},
\end{equation}
where $\mathbf{P}$ denotes the input point cloud and $\mathbf{R}$ is a rotation matrix in $\text{SO}(3)$, the group of all proper 3D rotation matrices.

Building upon Eq.~\ref{eq:invariance}, we propose Rotation-Invariant Handcrafted Features (RIHF) to encode geometric features (e.g., distances, angles) that are invariant to rotations. 
The input point cloud $\mathbf{P}$ is decomposed into $K$ local patches $\{\mathbf{P}_i\}_{i=1}^K$ via Farthest Point Sampling (FPS) for centroid selection and $k$-Nearest Neighbors (KNN) for neighborhood construction. 
During pretraining, a subset of patches is randomly masked as reconstruction targets.  
For each unmasked patch $\mathbf{P}_i$, RIHF enforces rotation invariance by:
\begin{equation}
\label{eq:rihf}
\text{RIHF}(\mathbf{P}_i) = \text{RIHF}(\mathbf{R}\mathbf{P}_i), \quad \forall \mathbf{R} \in \text{SO}(3).
\end{equation}

The extracted features are transformed into token embeddings that encode local geometric patterns and position embeddings that maintain spatial relationships.
These embeddings are then fed into the encoder to learn high-level geometric features.

However, the strict rotation invariance of RIHF creates a fundamental challenge for reconstruction tasks. 
Since RIHF produces identical features for $\mathbf{P}$ and $\mathbf{R}\mathbf{P}$, the decoder $g$ receives the same input features regardless of how the point cloud is rotated. 
This leads to an inherent ambiguity: the decoder cannot determine whether the original input was rotated or not.

To resolve this ambiguity, we redefine the reconstruction target as the aligned point cloud $\mathbf{P}_{\text{align}} = \text{Align}(\mathbf{R}\mathbf{P})$, which removes the rotational component through coordinate normalization. 
The reconstruction objective therefore becomes:
\begin{equation}
g\big(\text{RIHF}(\mathbf{R}\mathbf{P})\big) = g\big(\text{RIHF}(\mathbf{P})\big) \rightarrow \mathbf{P}_{\text{align}}.
\end{equation}

The key insight is that while the decoder cannot recover the original rotated instance $\mathbf{P}$ as the orientation information has been eliminated, it can reliably reconstruct the rotation-normalized geometry $\mathbf{P}_{\text{align}}$. 
This approach maintains rotation invariance in feature learning while providing a well-posed reconstruction target.

\subsection{HFBRI-MAE Architecture}

The proposed HFBRI-MAE framework extends the MAE architecture, as depicted in Fig.~\ref{fig:structure}. 
The input point cloud $\mathbf{P} \in \mathbb{R}^{N\times3}$ is first decomposed into $N_p$ patches. 
This process begins with FPS to select patch centroids $\{\mathbf{p}_i\}_{i=1}^{N_p}$ that maximize spatial coverage. 
Subsequently, KNN grouping is applied to form local neighborhoods. 
This two-stage approach ensures the preservation of both fine-grained local details and global structural context.

During the pretraining phase, we mask a high ratio subset of patches (50\%--80\%) by replacing them with learnable embedding tokens. 
The visible patches are first converted to RIHF, then converted to embeddings through designed embedding networks. 
Although adopting a fixed masking strategy facilitates training stability and implementation simplicity, it may limit the model’s adaptability to varying input structures. 
In contrast, adaptive masking—guided by factors such as geometric complexity or local feature density—offers a potential pathway to enhance the model’s flexibility and representation learning capacity. Investigating such dynamic masking strategies remains an important direction for future research.
The encoder $f$ processes these invariant embeddings to generate latent representations. 
A decoder $g$ then reconstructs the masked patches from these latent features, employing efficient self-attention mechanisms to minimize computational overhead.

For downstream finetuning, the masking mechanism is disabled to utilize the complete point cloud input. 
The pretrained encoder remains fixed while task-specific heads are appended for applications such as classification and segmentation. 

\subsection{Rotation-Invariant Handcrafted Features (RIHF)}
\begin{figure}[t]
    \centering
    \includegraphics[width=0.9\columnwidth]{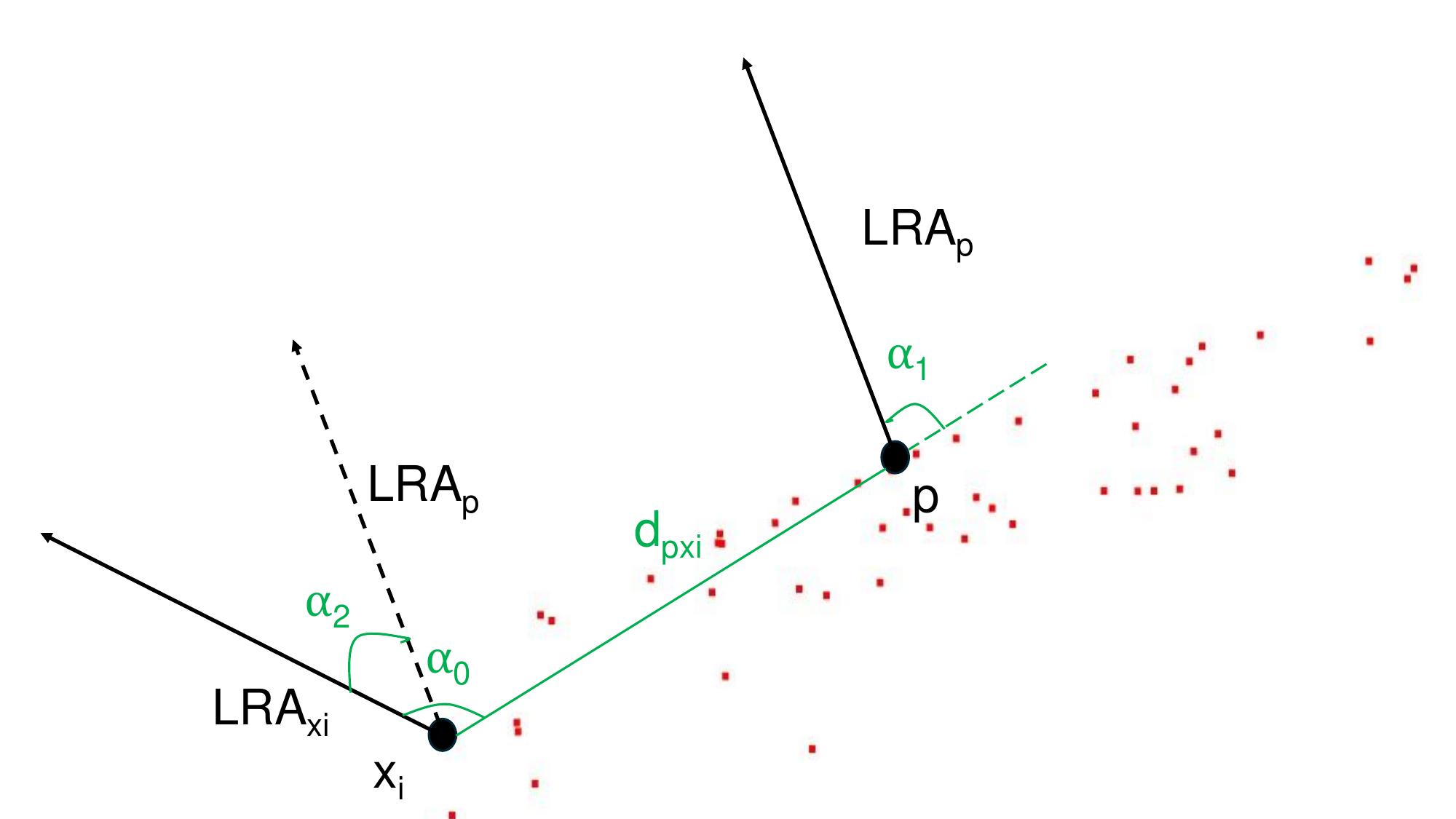}
    \caption{Visualization of distance features (\(d_{pxi}\)) and angle relationships with  reference point (\(\alpha_0, \alpha_1, \alpha_2\)) in RILF.}
    \label{fig:Local1}
\end{figure}
Our designed RIHF consists of two complementary components to achieve more complete rotation invariance: Rotation-Invariant Local Features (RILF) that capture fine-grained geometric patterns within local patches, and Rotation-Invariant Global Features (RIGF) that encode broader structural relationships across spatial regions.

\textbf{Rotation-Invariant Local Features (RILF).} We develop RILF, building upon RIConv++’s concepts~\cite{b8}, with all features designed within local patches. 
Our framework incorporates three feature types: distance feature, reference point angle features, and inter-neighbor angle features.
Surface normal vectors are commonly used as stable geometric references for constructing rotation-invariant features. 
However, many point cloud datasets lack explicit normal vectors, necessitating the development of alternative approaches.

We address this limitation by introducing the Local Reference Axis (LRA), derived from the smallest eigenvector of the local neighborhood's covariance matrix. 
Points are projected onto the tangent plane at the reference point $p$, and ordered clockwise, starting from the farthest point to maintain consistent point ordering relationships under arbitrary rotations.

\begin{figure}[t]
    \centering
    \includegraphics[width=0.9\columnwidth]{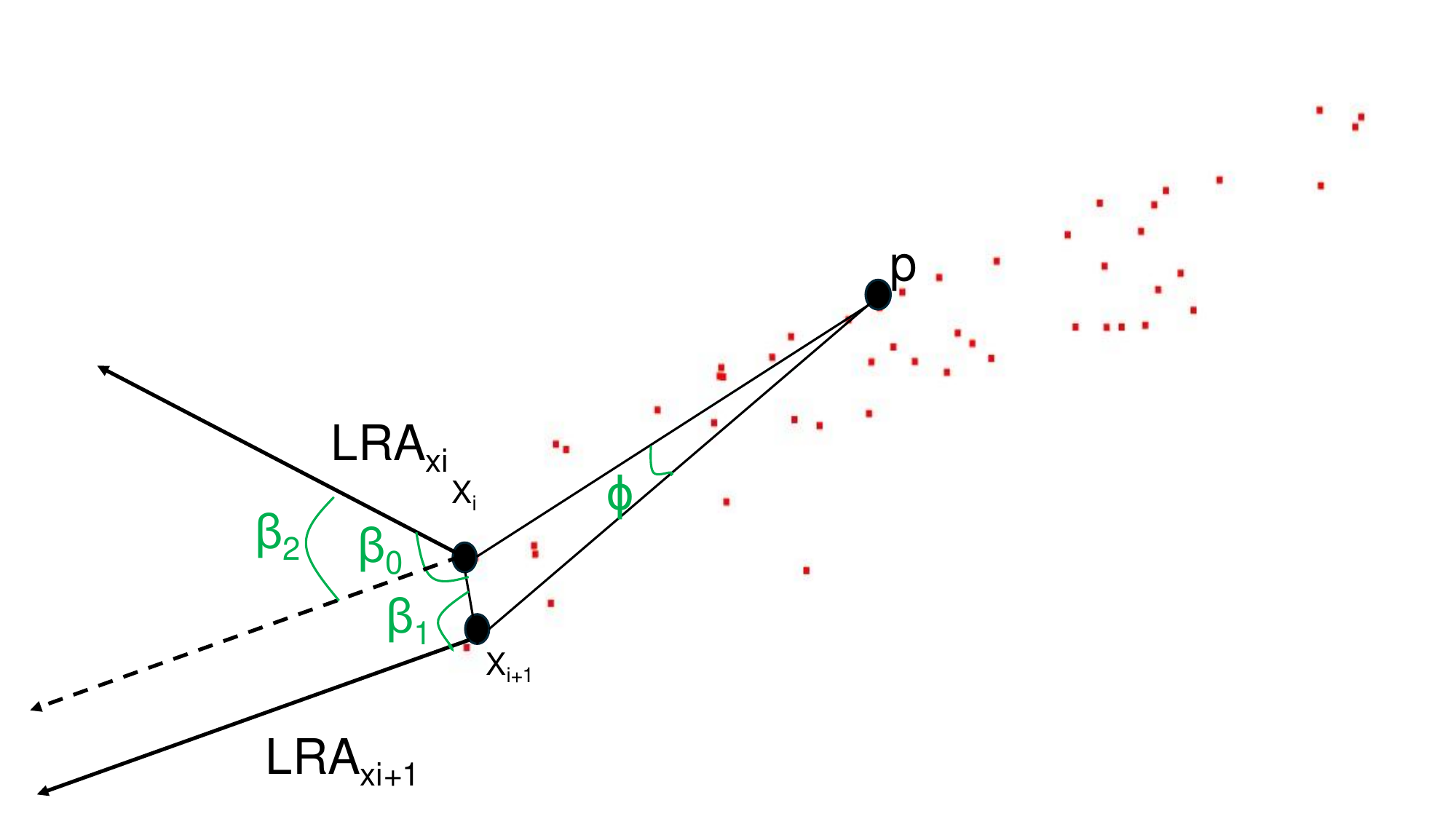}
    \caption{Visualization of inter-neighbor angle relationship (\(\phi, \beta_0, \beta_1, \beta_2\)) in RILF. }
    \label{fig:Local2}
\end{figure}

The distance feature ($d_{pxi}$) quantifies the spatial proximity between neighboring points and the reference point $p$, as illustrated in Fig.~\ref{fig:Local1}. 
The reference point angle features ($\alpha_0$, $\alpha_1$, $\alpha_2$) represent the angle relationships between the target point $x_i$ and the reference point $p$, including the LRA and connecting vectors.
The inter-neighbor angle features ($\phi$, $\beta_0$, $\beta_1$, $\beta_2$) encode the angle relationships between adjacent points $x_i$ and $x_{i+1}$, as shown in Fig.~\ref{fig:Local2}.

We construct the comprehensive RILF by concatenating these geometric features:
\begin{equation}
\text{RILF}(x_i) = [d_{pxi}, \alpha_0, \alpha_1, \alpha_2, \phi, \beta_0, \beta_1, \beta_2].
\end{equation}

\begin{figure}[t]
    \centering
    \includegraphics[width=0.9\columnwidth]{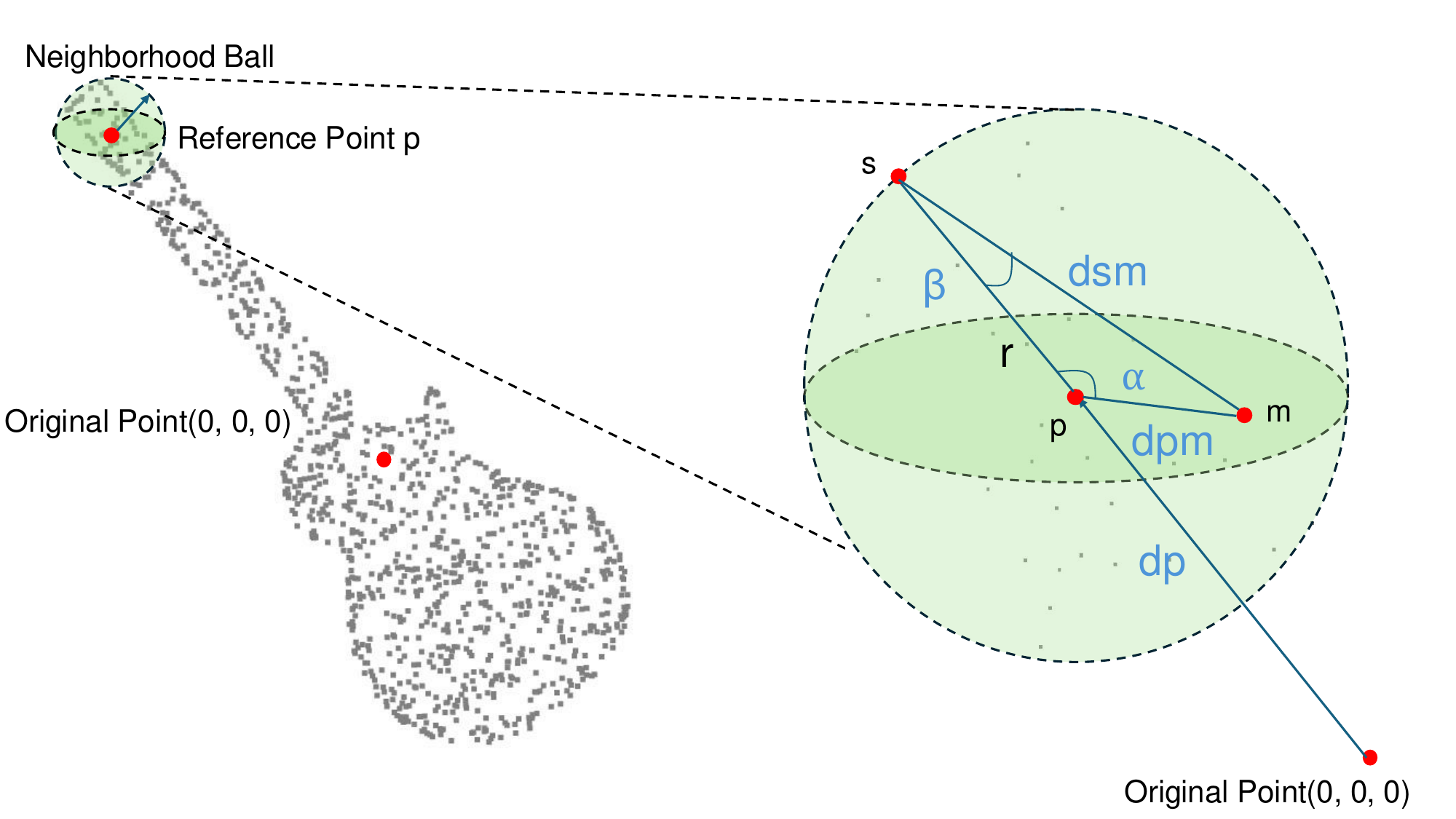}
    \caption{Visualisation of the RIGF (\(d_{p}, d_{pm}, d_{sm}, \alpha, \beta\)) construction process using the neighborhood ball, which is centered at the reference point $p$ with a radius $r$ defined by the distance to the farthest neighboring point.}
    \label{fig:global}
\end{figure}

\textbf{Rotation-Invariant Global Features (RIGF).} We propose RIGF to encode global spatial relationships using a neighborhood ball-based structure centered on the reference point $p$. 
Each ball corresponds exactly to the patch represented by the reference point, containing all the points within that patch. 
The ball’s radius $r$ is determined by the farthest point from the reference point within the patch~\cite{b30}.
As illustrated in Fig.~\ref{fig:global}, we define point $m$ as the centroid of the ball and point $s$ as the intersection between the ball boundary and the ray extending from the origin to reference point $p$. 
The RIGF incorporates three distance measurements ($d_p$, $d_{pm}$, $d_{sm}$) and two global angle features ($\alpha$, $\beta$). 
$d_p$ measures the origin-to-$p$ distance, $d_{pm}$ measures the $p$-to-$m$ distance, and $d_{sm}$ measures the $s$-to-$m$ distance. 
The angles $\alpha$ and $\beta$ quantify the relationships between vectors $\overrightarrow{p m}$, $\overrightarrow{p s}$, and $\overrightarrow{m s}$.
The complete RIGF representation is formulated as:
\begin{equation}
\text{RIGF}(pi) = [d_{p}, d_{pm}, d_{sm}, \alpha, \beta].
\end{equation}

\subsection{Encoder and Decoder}

To effectively process the RIHF, we enhance the MAE's encoder-decoder framework with three key components: embedding layers that transform RILF and RIGF into higher-dimensional representations, a transformer-based encoder for feature processing, and a decoder for point cloud reconstruction.

\textbf{Embedding.} The embedding process transforms RIHF into higher-dimensional representations required by the encoder-decoder architecture. 
This transformation operates through two parallel pathways.
RILF undergo mapping to token embeddings through a Multi-Layer Perceptron (MLP), which yield the representation in encoder space:
\begin{equation}
\mathbf{E}_\text{token} = \text{MLP}_\text{local}(\text{RILF}).
\end{equation}
A parallel transformation processes RIGF into position embeddings via another MLP, encoding the spatial relationships between patches:
\begin{equation}
\mathbf{E}_\text{position} = \text{MLP}_\text{global}(\text{RIGF}).
\end{equation}

\textbf{Encoder.} The encoder adopts a transformer-based architecture for processing rotation-invariant features extracted from point cloud patches. 
The architecture consists of 12 sequential transformer blocks, each implementing an 8-head self-attention mechanism. The multi-head self-attention operation is formulated as:
\begin{equation}
\text{Attention}(Q, K, V) = \text{softmax}\left(\frac{QK^\top}{\sqrt{d_k}}\right)V,
\end{equation}
where the attention mechanism operates on query ($Q$), key ($K$), and value ($V$) matrices, with $d_k$ denoting the dimension of keys. 
This formulation enables the model to capture geometric relationships across patches through learnable attention weights.

The encoder processes features progressively through its transformer blocks. 
The initial block receives the element-wise sum of token embeddings $\mathbf{E}_\text{token}$ and position embeddings $\mathbf{E}_\text{position}$ as input. 
Subsequent blocks integrate position embeddings with the processed features from their preceding blocks through residual connections. 
The output of the final block is a sequence of patch-level features, which can be directly fed into the decoder for reconstruction tasks.
For classification or other global tasks, we further apply max-pooling and average-pooling over the sequence, and sum their results to produce a compact global representation.

\textbf{Decoder.} The decoder reconstructs the masked point cloud patches by leveraging the encoder's output. 
Built on a transformer architecture with 4 decoding blocks, each utilizes an 8-head self-attention mechanism. 
The decoder processes $N_v$ visible tokens and $N_m$ masked tokens, where $N_m = N_p - N_v$ with $N_p$ denoting the total number of patches. 
The initial masked token, represented as a $d$-dimensional learnable vector, is duplicated $N_m$ times and concatenated with the visible tokens before being fed to the first decoding block. 
Through layer-wise self-attention operations, each block refines the embeddings by progressively capturing hierarchical geometric relationships. 
The final output is passed through a reconstruction head that predicts the coordinates of the masked points, thereby restoring the complete geometric structure of the point cloud.

\subsection{Objective of Pretraining and Finetuning}

\textbf{Objective of Pretraining.} In pretraining process, HFBRI-MAE aims to reconstruct aligned 3D coordinates within masked point cloud patches. 
A high masking ratio forces the model to infer missing geometric details from visible patches, focusing on meaningful geometric abstractions. 
The pretraining dataset consists of pre-aligned point clouds, which we directly use as $\mathbf{P}_{\text{align}}$ for the reconstruction target. 
The pretraining loss aims to minimize the discrepancy between predicted and aligned point clouds through the Chamfer Distance:
\begin{equation}
L_{\text{Chamfer}} = \sum_{x \in P_{\text{gt}}} \min_{y \in P_{\text{pred}}} \|x - y\|^2
+ \sum_{y \in P_{\text{pred}}} \min_{x \in P_{\text{gt}}} \|y - x\|^2,
\end{equation}
where \(P_{\text{gt}}\) and \(P_{\text{pred}}\) represent the ground truth and predicted aligned point clouds, respectively.

\textbf{Objective of Finetuning.} The pre-trained model undergoes finetuning to adapt its representations for downstream tasks. 
During this phase, the encoder's output is directly fed into a task-specific prediction head. 
The adaptation process for classification tasks optimizes cross-entropy loss.

\section{Experiments and Results}
\subsection{Experimental Setup}

\textbf{Rotation Settings.} To assess the rotation invariance capability of our model, we consider three rotation configurations of point clouds.
Aligned ($A$) represents unrotated data. 
Z-axis-constrained ($Z$) refers to rotations around the z-axis.
Random ($R$) applies uniformly sampled rotations in the 3D space, covering all possible orientations.
To evaluate rotation-invariant feature learning during pretraining, we train an SVM classifier on encoder features. 

Our evaluation system utilizes a consistent notation $X/Y$, where $X$ denotes the rotation applied during training (including both pretraining and subsequent finetuning or SVM training), and $Y$ represents the rotation during testing. 
From the nine possible rotation combinations, we identify five critical settings. 
These settings demonstrate comprehensive rotation invariance performance.
The $A/A$ configuration examines performance under global orientation consistency, while $Z/Z$ evaluates rotation robustness with z-axis constrained rotations. 
$A/R$ and $Z/R$ assess the generalization capability from structured to random rotations. 
The $R/R$ setting further demonstrates model robustness under fully random rotations. 
The experimental results of these five rotation settings will be presented in subsequent sections, while the complete analysis of all nine rotation settings is available in the supplementary material.

\textbf{Datasets.} Pretraining utilized ShapeNetCore55~\cite{b31}, a diverse dataset comprising 51,300 CAD models.
Finetuning was performed on ModelNet40~\cite{b32} and ScanObjectNN~\cite{b33}, covering synthetic and real-world scenarios, with all point clouds downsampled to 1024 points.

\textbf{Downstream Tasks.} The framework was evaluated on object classification, part segmentation, and few-shot learning. 
For classification, we used an SVM classifier on ModelNet40 when pretraining. 
During finetuning, we conducted separate training procedures for each dataset ModelNet40 and ScanObjectNN with their respective classification heads initialized from scratch.
Part segmentation was performed on ShapeNetPart using the category-level mean intersection-over-union (C-mIoU) metric. 
Few-shot learning experiments, following~\cite{b35}, were conducted on ModelNet40 with settings of ‘5-way 10-shot’ and ‘10-way 10-shot’.

\textbf{Parameter Setup.} Pretraining employed a batch size of 128, a learning rate of 0.001 with cosine decay and AdamW optimizer~\cite{b36} with a weight decay of 0.05~\cite{b37}, over 300 epochs.
A 60\% masking ratio was applied to 256 patches per point cloud, with each patch consisting of 64 points. 
Training was performed on a platform with an Intel Core i5-14600K CPU and an NVIDIA RTX 4090 GPU.

\subsection{Experimental Results}

\textbf{Object Classification on ModelNet40.} Table~\ref{tab:modelnet40} presents the classification accuracy across the defined rotation settings. 
Non-RI methods exhibit significant performance degradation under $Z/R$ and $R/R$ settings, reflecting their limited robustness to rotational variations. 
RI methods, such as RIConv++ and PaRot, deliver consistent performance but are surpassed by HFBRI-MAE.

HFBRI-MAE achieves classification accuracies exceeding 91.3\% across all rotation settings. 
Notably, it achieves state-of-the-art performance in all settings containing rotated data, underscoring the model’s effectiveness in handling diverse rotational scenarios. 
Furthermore, the superiority of the finetuned model over the pretrained configuration highlights the significant benefits of task-specific optimization in enhancing performance.

\begin{table}[t]
\centering
\caption{Classification performance on ModelNet40 ($A/A$, $A/R$, $Z/Z$, $Z/R$, $R/R$). ``RI" denotes whether the method is rotation invariance. ``Pretrain" represents SVM classifier results.}
\label{tab:modelnet40}
\begin{tabular}{l|c|c|c|c|c|c}
\toprule
\textbf{Methods} & \textbf{RI} & \textbf{A/A} & \textbf{A/R} & \textbf{Z/Z} & \textbf{Z/R} & \textbf{R/R} \\ 
\midrule
PointNet~\cite{b1} &  & 89.2 & 18.2 & 85.9 & 70.2 & 83.1 \\ 
PointNet++~\cite{b2} &  & 90.3 & 22.8 & 89.3 & 72.3 & 85.0 \\ 
DGCNN~\cite{b38} &  & 91.7 & 23.1 & 91.2 & 76.4 & 86.1 \\ \midrule 
RIConv~\cite{b7} & \checkmark & 86.2 & 85.9 & 86.5 & 86.4 & 86.3 \\ 
RI-Framework~\cite{b30} & \checkmark & 89.3 & 89.3 & 89.4 & 89.4 & 89.3 \\ 
Li et al.~\cite{b17} & \checkmark & 89.8 & 90.0 & 90.2 & 90.2 & 90.2 \\ 
OrientedMP~\cite{b43} & \checkmark & 88.8 & 88.7 & 88.4 & 88.4 & 88.9 \\ 
PaRot~\cite{b12} & \checkmark & 91.1 & 90.9 & 90.9 & 91.0 & 90.8 \\ 
RIConv++~\cite{b8} & \checkmark & 91.2 & 90.8 & 91.3 & 91.2 & 91.2 \\ \midrule
\multicolumn{7}{c}{\textbf{SSL Methods}} \\ \midrule
PointMAE~\cite{b25} &  & 91.8 & 24.2 & 87.2 & 46.7 & 81.5 \\ 
PointM2AE~\cite{b26} &  & \textbf{92.6} & 24.3 & 90.1 & 39.6 & 84.4 \\
MaskSurf~\cite{b44}&  & 92.4 & 35.3 & 90.5 & 56.3 & 85.9 \\
PointGPT~\cite{b45}&  & \textbf{92.6} & 40.5 & 90.5 & 45.2 & 86.2 \\
RI-MAE~\cite{b28} & \checkmark & 89.5 & 89.6 & 89.4 & 89.3 & 89.7 \\
MaskLRF~\cite{b27} & \checkmark & 91.3 & 90.6 & 91.0 & 91.2 & 91.0 \\ \midrule
HFBRI-MAE (Pretrain) & \checkmark & 89.7 & 89.8 & 89.9 & 89.6 & 89.5 \\ 
HFBRI-MAE (Finetune) & \checkmark & 91.3 & \textbf{91.4} & \textbf{91.7} & \textbf{91.3} & \textbf{91.5} \\ 

\bottomrule
\end{tabular}
\end{table}

\begin{table}[t]
\centering
\caption{Classification performance on ScanObjectNN-BG ($A/A$, $A/R$, $Z/Z$, $Z/R$, $R/R$). ``RI" denotes whether the method is rotation invariance. ``-BG" denotes the data with background.}
\label{tab:scanobjectnn}
\begin{tabular}{l|c|c|c|c|c|c}
\toprule
\textbf{Methods} & \textbf{RI} & \textbf{A/A} & \textbf{A/R} & \textbf{Z/Z} & \textbf{Z/R} & \textbf{R/R} \\ 
\midrule
PointNet~\cite{b1} &  & 73.3 & 13.2 & 72.9 & 71.3 & 70.2 \\ 
PointNet++~\cite{b2} &  & 82.3 & 21.2 & 81.1 & 68.5 & 76.3 \\ 
DGCNN~\cite{b38} &  & 82.8 & 31.1 & 81.2 & 69.8 & 80.1 \\ 
\midrule
RIConv~\cite{b7} & \checkmark & 78.4 & 77.9 & 78.5 & 78.8 & 77.8 \\ 
RI-Framework~\cite{b30} & \checkmark & 80.5 & 81.5 & 79.4 & 79.8 & 79.9 \\ 
Li et al.~\cite{b17} & \checkmark & 78.8 & 80.0 & 79.2 & 79.3 & 79.6 \\ 
RIConv++~\cite{b8} & \checkmark & 89.7 & 90.0 & 89.9 & 89.1 & 89.7 \\ 
OrientedMP~\cite{b43} & \checkmark & 76.8 & 78.1 & 77.6 & 76.7 & 77.2 \\ 
PaRot~\cite{b12} & \checkmark & 88.2 & 88.5 & 88.1 & 88.0 & 88.0 \\ 
\midrule
\multicolumn{7}{c}{\textbf{SSL Methods}} \\ \midrule
PointMAE~\cite{b25} &  & 90.0 & 26.5 & 85.4 & 42.3 & 86.1 \\ 
PointM2AE~\cite{b26} &  & 91.2 & 32.8 & 86.3 & 46.3 & 86.3 \\ 
MaskSurf~\cite{b44}&  & 91.2 & 29.0 & 86.4 & 39.6 & 89.2 \\ 

PointGPT~\cite{b45}&  & \textbf{91.6} & 31.5 & 84.8 & 45.2 & 85.4 \\ 
\midrule
HFBRI-MAE & \checkmark & 90.1 & \textbf{90.5} & \textbf{90.0} & \textbf{90.3} & \textbf{90.0} \\ 
\bottomrule
\end{tabular}
\end{table}

\textbf{Object Classification on Real-World Datasets.} 
The ScanObjectNN-BG dataset presents significant challenges with noisy and occluded point clouds.  
As shown in Table~\ref{tab:scanobjectnn}, non-RI methods experience substantial accuracy losses in $A/R$, $Z/R$, and $R/R$ settings, reflecting their inability to handle complex real-world scenarios.  
While RI methods exhibit more stable performance, they are consistently outperformed by HFBRI-MAE.  
These results demonstrate HFBRI-MAE’s robustness under diverse rotations and environmental noise.

\begin{table*}[t]
\centering
\caption{Segmentation per class results and averaged class mIoU on ShapeNetPart dataset under Z/R, where C-mIoU stands for averaged mIoU of 16 classes.}
\label{tab:segmentation}
{\setlength{\tabcolsep}{3pt}
\begin{tabular}{l|c|c|cccccccccccccccc}
\toprule
\textbf{Methods} & \textbf{RI} & \textbf{C-mIoU} 
                & \textbf{aero} & \textbf{bag} & \textbf{cap} & \textbf{car} & \textbf{chair} & \textbf{earph.} & \textbf{guitar} & \textbf{knife} & \textbf{lamp} & \textbf{laptop} & \textbf{motor} & \textbf{mug} & \textbf{pistol} & \textbf{rocket} & \textbf{skate} & \textbf{table} \\ 
\midrule
PointNet~\cite{b1} &  & 37.8 
                & 40.4 & 48.1 & 46.3 & 24.5 & 45.1 & 39.4 & 29.2 & 42.6 & 52.7 & 36.7 & 21.2 & 55.0 & 29.7 & 26.6 & 32.1 & 35.8 \\
PointNet++~\cite{b2} &  & 48.3 
                & 51.3 & 66.0 & 50.8 & 25.2 & 66.7 & 27.7 & 29.7 & 65.6  & 59.7 & 70.1 & 17.2 & 67.3 & 49.9 & 23.4 & 43.8 & 57.6 \\
DGCNN~\cite{b38} &  & 37.4 
                & 37.0 & 50.2 & 38.5 & 24.1 & 43.9 & 32.3 & 23.7 & 48.6  & 54.8 & 28.7 & 17.8 & 74.4 & 25.2 & 24.1 & 43.1 & 32.3 \\ 
\midrule
Li et al.~\cite{b17} & \checkmark & 74.1 
                & 81.9 & 58.2 & 77.0 & 71.8 & \textbf{89.6} & 64.2 & 89.1 & \textbf{85.9} & 80.7 & \textbf{84.7} & 46.8 & 89.1 & 73.2 & 45.6 & 66.5 & \textbf{81.0} \\
RIConv~\cite{b7} & \checkmark & 75.3 
                & 80.6 & 80.0 & 70.8 & 68.8 & 86.8 & 70.3 & 87.3 & 84.7 & 77.8 & 80.6 & 57.4 & 91.2 & 71.5 & 52.3 & 66.5 & 78.4 \\
RIConv++~\cite{b8} & \checkmark & 78.9 & 
            82.1 & 80.0 & 86.7 & 76.5&  89.3 & 64.5&  89.8&  83.8&  81.9&  81.3&  65.0 & 93.3&  78.1&  53.7 & 72.7 & 80.3\\
PaRot~\cite{b12} & \checkmark & 79.2 
                & \textbf{82.7} & 79.2 & 82.3 & \textbf{75.3} & 89.4 & \textbf{73.9} & 89.1 & 85.6  & 81.0 & 79.5 & 65.3 & 93.9 & \textbf{79.2} & 55.0 & 72.4 & 79.5 \\ 
\midrule
\multicolumn{19}{c}{\textbf{SSL Methods}} \\ \midrule
PointMAE~\cite{b25} & & 34.0& 36.6& 31.8& 38.5& 31.1& 33.1& 31.9& 37.2& 37.5& 39.0& 35.0& 25.7& 43.4& 32.4& 29.4& 31.9& 29.9 \\
PointM2AE~\cite{b26} && 39.6& 41.1& 39.3& 49.5& 34.6& 44.7& 39.2& 36.0& 48.6& 38.4& 33.0& 33.5& 49.9& 37.7& 36.2& 41.4&30.6 \\ \midrule
HFBRI-MAE & \checkmark & \textbf{79.5} 
                & 82.6 & \textbf{82.5} & \textbf{87.6} & 72.0 & 88.4 & 72.7 & \textbf{90.9} & 83.5  & \textbf{83.1} & 80.8 & \textbf{66.0} & \textbf{94.2} & 76.5 & \textbf{57.4} & \textbf{73.7} & 80.0 \\ 
\bottomrule
\end{tabular}
}
\end{table*}

\textbf{Part Segmentation on ShapeNet.} Table~\ref{tab:segmentation} demonstrates HFBRI-MAE's superior performance in part segmentation on the ShapeNetPart dataset under $Z/R$ rotation conditions. 
The model attains a C-mIoU of 79.5\%, surpassing existing rotation-invariant approaches. 
Its effectiveness manifests particularly in complex categories such as \textit{guitar} and \textit{lamp}, outperforming other methods across different categories. 
These results validate the model's capacity to maintain detailed semantic feature recognition under various rotational transformations.

\begin{figure}[t]
    \centering
    \includegraphics[width=\columnwidth]{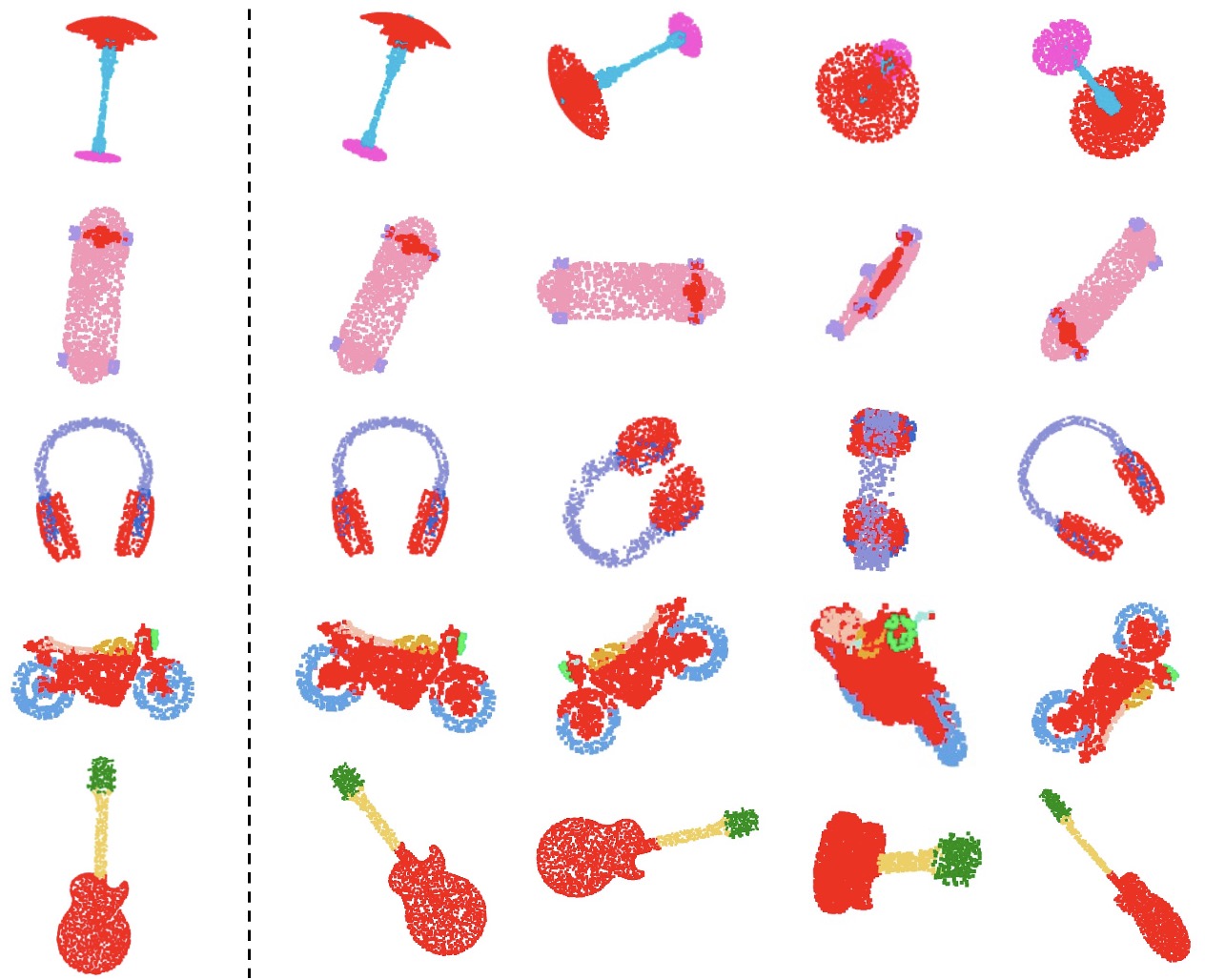}
    \caption{Visualization of part segmentation results on the ShapeNetPart dataset under Z/R.}
    \label{fig:visualization}
\end{figure}

Fig.~\ref{fig:visualization} provides qualitative results of segmentation on several challenging categories. 

\begin{table}[t]
\centering
\caption{Performance of 5-way 10-shot few-shot on ModelNetFewShot dataset.}
\label{tab:5_10fewshot}
\begin{tabular}{l|c|c|c|c|c|c}
\toprule
\textbf{Methods} & \textbf{RI} & \textbf{A/A} & \textbf{A/R} & \textbf{Z/Z} & \textbf{Z/R} & \textbf{R/R} \\ 
\midrule
PaRI-Conv~\cite{b11} & \checkmark & 89.9 & 90.7 & 90.8 & 89.0 & 90.5 \\ 
RIConv++~\cite{b8} & \checkmark & 87.3 & 87.9 & 87.6 & 87.0 & 87.5 \\ 
PaRot~\cite{b12} & \checkmark & 46.9 & 49.5 & 48.7 & 49.5 & 49.0 \\ 
\midrule
\multicolumn{7}{c}{\textbf{SSL Methods}} \\ \midrule
PointMAE~\cite{b25} &  & 96.3 & 52.6 & 85.2 & 56.7 & 87.7 \\ 
PointM2AE~\cite{b26} &  & \textbf{96.8} & 58.1 & 86.3 & 61.4 & 89.1 \\ 
MaskSurf~\cite{b44} &  & 96.5 & 50.9 & 84.2 & 55.5 & 90.9 \\ 
PointGPT~\cite{b45} &  & \textbf{96.8} & 48.2 & 88.3 & 59.0 & 84.3 \\ 
RI-MAE~\cite{b28} & \checkmark & 92.2 & 92.3 & 93.1 & 92.7 & 92.5 \\ 
MaskLRF~\cite{b27} & \checkmark & 93.5 & 93.6 & 93.2 & 93.7 & \textbf{93.8} \\ \midrule
HFBRI-MAE & \checkmark & 93.9 & \textbf{94.4} & \textbf{93.7} & \textbf{93.8} & 93.5 \\ 
\bottomrule
\end{tabular}
\end{table}
The visualization highlights HFBRI-MAE’s capability to preserve semantic consistency across object parts, accurately segmenting fine details like the \textit{lamp shade} and \textit{guitar neck}. 
These results underscore the robustness and effectiveness of the proposed method in achieving consistent segmentation performance under rotational transformations.

\begin{table}[t]
\centering
\caption{Performance of 10-way 10-shot few-shot on ModelNetFewShot dataset.}
\label{tab:10_10fewshot}
\begin{tabular}{l|c|c|c|c|c|c}
\toprule
\textbf{Methods} & \textbf{RI} & \textbf{A/A} & \textbf{A/R} & \textbf{Z/Z} & \textbf{Z/R} & \textbf{R/R} \\ 
\midrule
PaRI-Conv~\cite{b11} & \checkmark & 84.6 & 84.6 & 85.2 & 84.5 & 84.0 \\ 
RIConv++~\cite{b8} & \checkmark & 80.0 & 80.3 & 80.9 & 80.4 & 80.8 \\ 
PaRot~\cite{b12} & \checkmark & 50.9 & 48.3 & 47.8 & 49.5 & 47.7 \\ \midrule
\multicolumn{7}{c}{\textbf{SSL Methods}} \\ 
\midrule
PointMAE~\cite{b25} &  & 92.6 & 33.6 & 79.6 & 42.5 & 80.2 \\ 
PointM2AE~\cite{b26} &  & 92.3 & 33.2 & 81.8 & 57.2 & 83.6 \\ 
MaskSurf~\cite{b44} &  & \textbf{93.0} & 33.0 & 78.1 & 62.7 & 83.3 \\ 
PointGPT~\cite{b45} &  & 92.6 & 30.2 & 80.9 & 60.6 & 79.6 \\ 
RI-MAE~\cite{b28} & \checkmark & 89.3 & 89.3 & 89.5 & 89.7 & 89.3 \\ 
MaskLRF~\cite{b27} & \checkmark & 89.2 & 89.0 & 89.3 & 89.2 & 89.5 \\ \midrule
HFBRI-MAE & \checkmark & 89.8 & \textbf{90.7} & \textbf{90.0} & \textbf{90.4} & \textbf{89.7} \\ 
\bottomrule
\end{tabular}
\end{table}
\textbf{Few-shot Classification on ModelNet40.} Table~\ref{tab:5_10fewshot} and Table~\ref{tab:10_10fewshot} summarize the results for the ‘5-way 10-shot’ and ‘10-way 10-shot’ settings on the ModelNetFewShot dataset. 
HFBRI-MAE consistently demonstrates superior performance across all rotation settings, highlighting its robust rotation-invariant capabilities.
In both experiment settings, HFBRI-MAE outperforms competing methods, showcasing strong generalization and adaptability under various rotational scenarios. 

\subsection{Ablation Experiment}

\textbf{Mask Ratio Analysis.} The mask ratio significantly impacts the model’s pretraining effectiveness by balancing visible and masked information. 
As shown in Table~\ref{tab:mask}, a ratio of 0.6 achieves the best classification accuracy of 89.6\% on ModelNet40. 
Lower ratios retain too much visible information, limiting feature abstraction, while higher ratios overly challenge the model, reducing performance. 
This underscores the importance of a balanced masking strategy.

\begin{table}[t]
\centering
\caption{Performance of classification on ModelNet40 dataset with different mask ratio (Pretrain + SVM).}
\label{tab:mask}
\begin{tabular}{c|c}
\toprule
\textbf{Mask Ratio} & \textbf{Average Acc.} \\ 
\midrule
0.3                 & 84.6         \\ 
0.5                 & 87.0                 \\ 
0.8                 & 86.7            \\ 
\midrule
0.6                 & \textbf{89.7} \\ 
\bottomrule
\end{tabular}

\end{table}

\begin{table}[t]
\centering
\caption{Performance of classification on ModelNet40 dataset with types of RILF.}
\label{tab:lrif_types}
\begin{tabular}{l|c}
\toprule
\textbf{Types of RILF} & \textbf{Average Acc.}  \\ 
\midrule
Without distance features (\(d_{pxi}\)) &88.7                   \\ 
Without reference point angels (\(\alpha_0, \alpha_1, \alpha_2\))&88.1        \\ 
Without inter-neighbor angels (\(\phi, \beta_0, \beta_1, \beta_2\))  &  91.1        \\ 
\midrule
All                & \textbf{91.4} \\ 
\bottomrule
\end{tabular}

\end{table}

\textbf{Types of RILF.} Table~\ref{tab:lrif_types} shows the effect of different RILF components on classification performance. 
Removing distance features ($d_{px_i}$) or reference point angles ($\alpha_0, \alpha_1, \alpha_2$) significantly reduces accuracy, highlighting their importance. 
In contrast, removing inter-neighbor angles ($\phi, \beta_0, \beta_1, \beta_2$) has a less pronounced impact, with accuracy dropping to 91.1\%. 
The full feature set achieves the best performance at 91.4\%, confirming the value of all features combined.

\section{Conclusion}
This paper introduced HFBRI-MAE, a novel masked autoencoder for rotation-invariant analysis of 3D point clouds. 
By integrating rotation-invariant handcrafted features, the model ensures robust and consistent feature extraction across arbitrary orientations, addressing a critical limitation of conventional MAE-based approaches.
Unlike conventional MAE-based approaches, HFBRI-MAE employs aligned point clouds as reconstruction targets, addressing the loss of rotational information inherent in handcrafted features. 
Extensive experiments on benchmark datasets demonstrate the superiority of the proposed framework in object classification, segmentation, and few-shot learning tasks. 
HFBRI-MAE outperforms state-of-the-art methods under various rotation settings, showcasing its adaptability and effectiveness in both synthetic and real-world scenarios.

\newpage

\newpage
\appendix

\appendixsection{RIHF Details}

\textbf{RILF.} RILF consists of three types of features that encode invariant geometric relationships within each patch.

The first type comprises distance-based features, specifically the Euclidean distance $d_{pxi} = | x_i - p |$ between reference point $p$ and its $i$-th neighbor $x_i$, which captures essential local spatial relationships.

The second type captures angular relationships with respect to the reference point through three critical angles: the polar angle $\alpha_0 = \angle (LRA_{x_i}, \overrightarrow{x_i p})$ between the Local Reference Axis (LRA) of $x_i$ and the vector from $x_i$ to $p$; the azimuthal angle $\alpha_1 = \angle (LRA_p, \overrightarrow{x_i p})$ between the reference point's LRA and the vector $\overrightarrow{x_i p}$; and a signed angle $\alpha_2 = S_a \cdot \angle (LRA_{x_i}, LRA_p)$ between the LRAs of $p$ and $x_i$, where $S_a$ preserves rotational directionality.

The third type encompasses geometric relationships among neighboring points, including the angle $\phi = \angle (\overrightarrow{x_{i+1} p}, \overrightarrow{x_i p})$ between consecutive neighbor vectors, the azimuthal angle $\beta_0 = \angle (LRA_{x_i}, \overrightarrow{x_i x_{i+1}})$ between a neighbor's LRA and the vector to the next neighbor, the polar angle $\beta_1 = \angle (LRA_{x_{i+1}}, \overrightarrow{x_i x_{i+1}})$ and a signed angle $\beta_2 = S_b \cdot \angle (LRA_{x_i}, LRA_{x_{i+1}})$ between consecutive LRAs.

\textbf{RIGF.} RIGF contains distance-based features and angular features. 

The distance-based features include three key measurements: the distance to origin $d_{pi} = |p_i|$ for each reference point $p_i$, providing global positioning; the distance to geometric median $d_{pmi} = |p_i - m_i|$ where $m_i$ represents the geometric median of the neighborhood ball, capturing local density information; and the distance $d_{smi} = |s_i - m_i|$ between the boundary point $s_i$ and the geometric median, encoding the spatial extent of local neighborhoods.

The angular features comprise two principal measurements: $\alpha_i = \angle (\overrightarrow{p_i m_i}, \overrightarrow{s_i m_i})$, measuring the angle between the reference point and boundary point vectors from the geometric median, and $\beta_i = \angle (\overrightarrow{p_i s_i}, \overrightarrow{m_i s_i})$, capturing the angular relationship between the boundary point and geometric median vectors from the reference point perspective. These features together provide a complete characterization of the point cloud's geometry while maintaining rotation invariance at both local and global scales.\\

\appendixsection{Embedding Details}

\begin{figure}[t]
    \centering
    \includegraphics[width=\columnwidth]{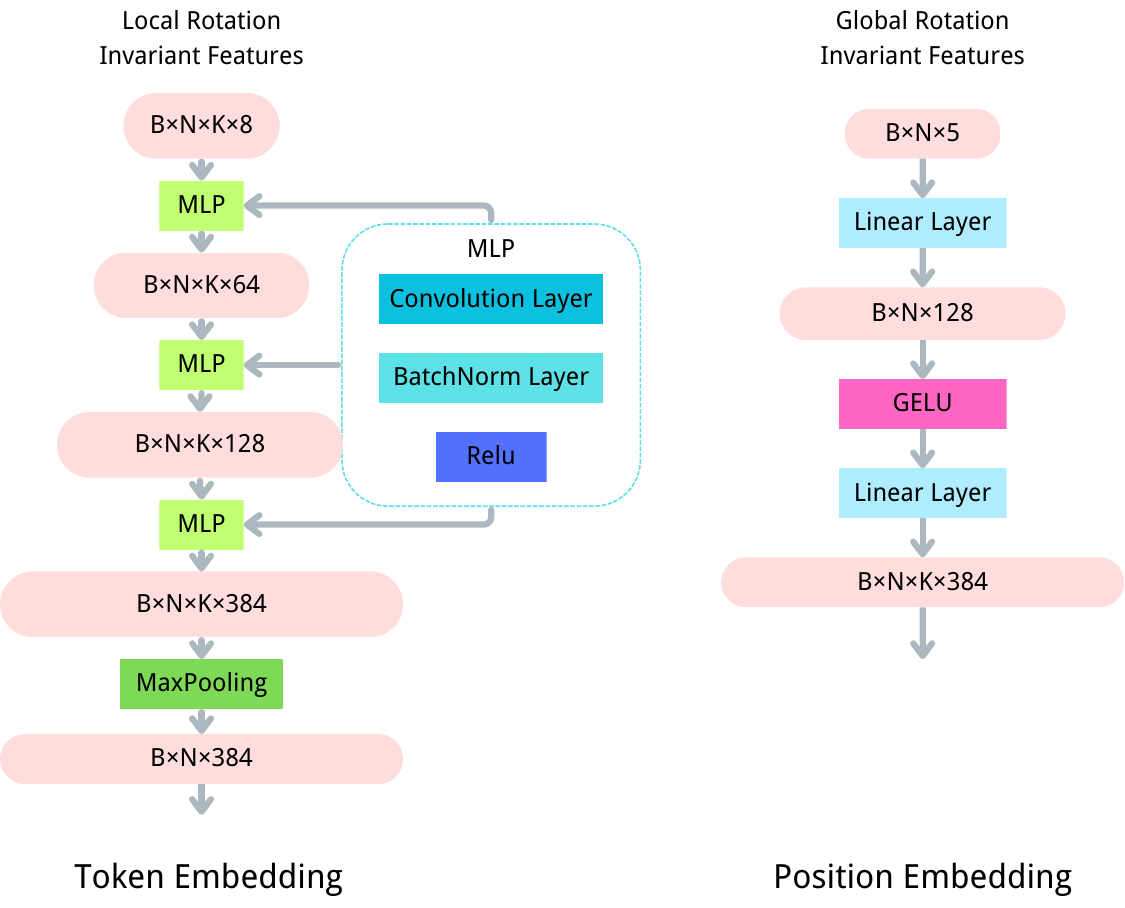}
    \caption{Embedding architecture overview. The left branch shows the token embedding pathway, which processes local rotation-invariant features through multiple MLP blocks (consisting of convolution, batch normalization, and ReLU activation) and a final max pooling operation. The right branch illustrates the position embedding pathway, which transforms global rotation-invariant features through a simpler architecture of linear layers and GELU activation. Both pathways produce embeddings of compatible dimensions (B×N×384) for subsequent transformer processing.}
    \label{fig:embedding}
\end{figure}

\textbf{Token Embedding.} As shown in Fig.~\ref{fig:embedding}, the token embedding pathway processes eight-dimensional local rotation-invariant features with an initial dimension of B×N×K×8, where B represents the batch size, N indicates the number of patches, and K denotes the number of points in each patch. This pathway implements a sophisticated multi-layer perceptron (MLP) structure, where each MLP block comprises three essential components: a convolution layer for feature transformation, batch normalization for training stability, and ReLU activation for non-linearity.

The network architecture progressively expands the feature dimensionality through three sequential MLP blocks. The first block transforms the initial 8-dimensional features to 64 dimensions (B×N×K×64), followed by a second block that further expands to 128 dimensions (B×N×K×128). The final MLP block produces a 384-dimensional representation (B×N×K×384). To consolidate the point-wise features within each patch, a max pooling operation is applied across the K dimension, resulting in the final token embedding of dimension B×N×384.

\textbf{Position Embedding.} The position embedding pathway processes five-dimensional global rotation-invariant features, starting with an input dimension of B×N×5. In contrast to the token embedding pathway, this architecture employs a more streamlined structure optimized for global spatial information. The pathway begins with a linear layer that expands the feature dimension from 5 to 128, producing an intermediate representation of B×N×128.
This intermediate representation then passes through a GELU activation function, introducing non-linearity while maintaining smooth gradients. A final linear layer transforms the features to match the token embedding dimension, producing a position embedding of B×N×K×384. This simpler architecture effectively captures global spatial relationships while maintaining dimensional compatibility with the token embeddings for subsequent transformer processing.

\appendixsection{Network Architecture for Downstream Tasks}

\begin{figure}[t]
    \centering
    \includegraphics[width=0.6\columnwidth]{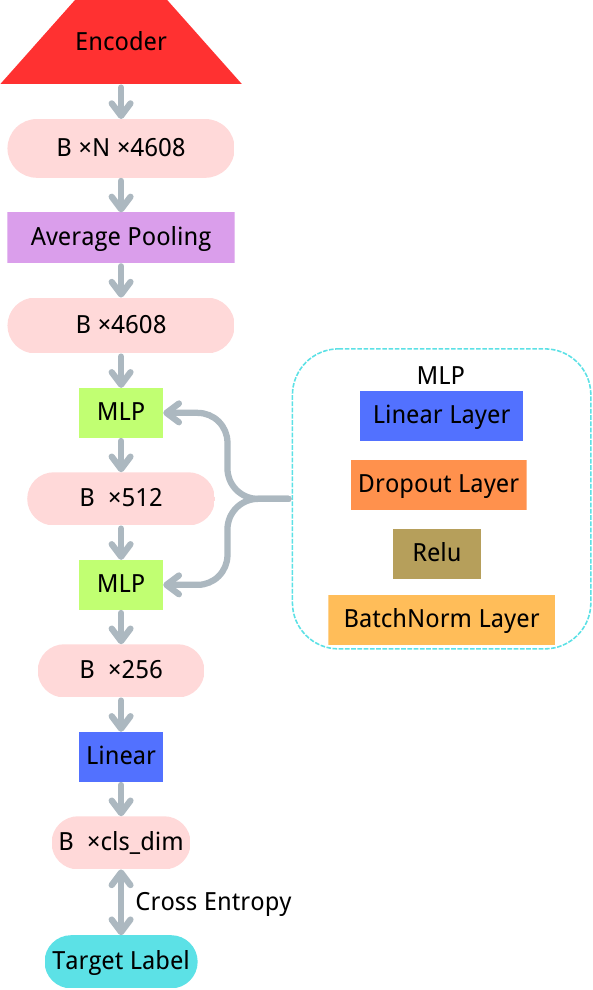}
    \caption{Classification head architecture. The network processes encoder outputs through average pooling and progressive dimensionality reduction via MLP blocks, culminating in class-specific predictions.}
    \label{fig:classification}
\end{figure}

\textbf{Classification Head.} As illustrated in Fig.~\ref{fig:classification}, our classification head architecture efficiently processes the encoder's output features ($B \times N \times 4608$) through a series of carefully designed transformations. The processing begins with an average pooling operation across the patch dimension, consolidating patch-wise features into a global $B \times 4608$ representation. This pooled feature vector then passes through two sequential MLP blocks, each comprising a linear layer, dropout, ReLU activation, and batch normalization. These MLP blocks progressively reduce the feature dimensionality from 4608 to 512, and then to 256, before a final linear layer maps the features to the target classification dimension ($B \times \text{cls\_dim}$). This architecture effectively distills the essential geometric information from the encoder's output while maintaining robust classification performance.

\begin{figure}[!t]
    \centering
    \includegraphics[width=0.7\columnwidth]{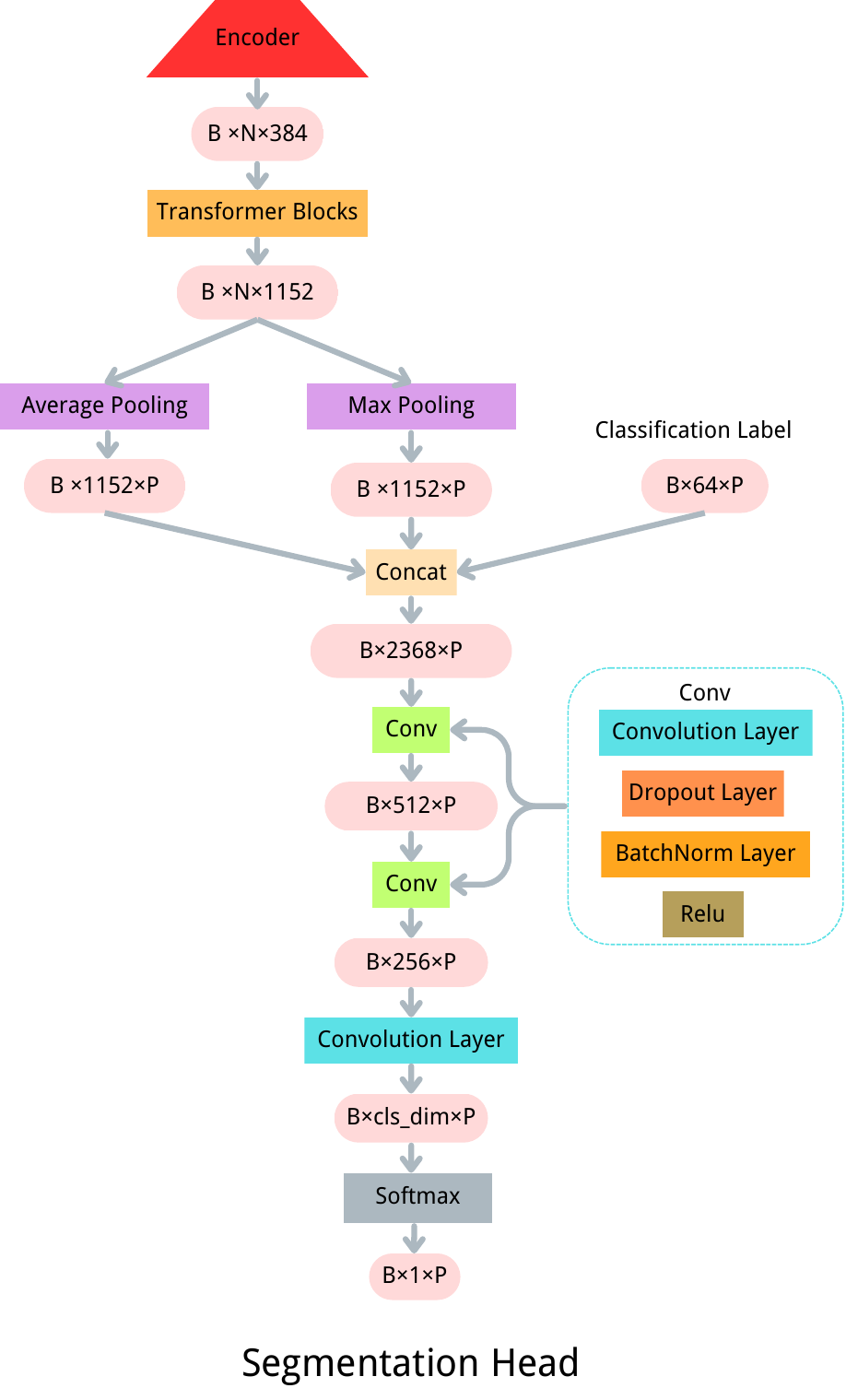}
    \caption{Architecture of the segmentation head. The diagram illustrates the flow from encoder features through parallel pooling branches and classification label integration, followed by progressive feature refinement via convolutional blocks, culminating in point-wise segmentation predictions.}
    \label{fig:segmentation}
\end{figure}

\textbf{Segmentation Head.}
As shown in Fig.~\ref{fig:segmentation}, our segmentation head processes the encoder's multi-scale features through a carefully designed architecture that combines global contextual information with point-wise predictions. The network begins by processing the encoder output of dimension $B \times N \times 384$ through transformer blocks, which expand the feature dimension to $B \times N \times 1152$, providing richer geometric representations.

The expanded features then flow through three parallel branches. Two branches perform average and max pooling operations respectively, both producing features of dimension $B \times 1152 \times P$, where $P$ represents the number of points. The third branch processes classification labels to generate point-wise label features of dimension $B \times 64 \times P$. These three streams are concatenated to form a comprehensive feature representation of dimension $B \times 2368 \times P$, effectively combining global contextual information with local geometric details.

This concatenated feature tensor then undergoes progressive refinement through a series of convolutional blocks. Each convolutional block consists of a convolution layer, followed by batch normalization, ReLU activation, and dropout for regularization. The first convolution reduces the feature dimension to $B \times 512 \times P$, followed by a second convolution that further reduces it to $B \times 256 \times P$. A final convolution layer maps the features to the target segmentation dimension $B \times \text{cls\_dim} \times P$, where $\text{cls\_dim}$ represents the number of segmentation classes. The architecture concludes with a softmax operation that produces point-wise segmentation probabilities of dimension $B \times 1 \times P$. 

\appendixsection{More Ablation Experiments}

\textbf{Completed Classfication Results. }Tables~\ref{tab:modelnet401}, \ref{tab:scanobjectnn1}, and \ref{tab:omniobject3d1} present the classification results of various methods across nine rotation settings on the ModelNet40, ScanObjectNN, and OmniObject3D datasets, respectively. These datasets provide a comprehensive evaluation of the rotation-invariant capabilities of the proposed HFBRI-MAE framework.

\textbf{Patch Count and Point Density.} The number of patches and points per patch directly influences the model’s ability to capture local and global geometric details.
Table~\ref{tab:knn} shows that the configuration with 256 patches and 64 points per patch achieves the best performance, with an accuracy of 89.6\% on ModelNet40. 
Smaller patch counts with fewer points result in insufficient feature coverage, while excessive patch numbers lead to redundancy and decreased generalization. 
This demonstrates the importance of selecting an appropriate balance between patch granularity and local detail representation.
\begin{table}[!t]
\centering
\caption{Performance of classification on ModelNet40 dataset with different numbers of patches and points per patch (Pretrain + SVM).}
\label{tab:knn}
\begin{tabular}{cc|c}
\toprule
\textbf{Number of Patches} & \textbf{Points Per Patch} & \textbf{Average Acc.} \\ 
\midrule
128                        & 16                        & 83.0                  \\ 
128                        & 64                        & 86.3                  \\ 
256                        & 16                        & 85.4                  \\ 
256                        & 32                        & 88.2                  \\ 
512                        & 32                        & 87.1                  \\ 
512                        & 64                        & 84.7                  \\ 
\midrule
256                        & 64                        & \textbf{89.6}         \\ 
\bottomrule
\end{tabular}
\end{table}

\begin{table}[!t]
\centering
\caption{Performance of classification on ModelNet40 dataset with different finetune strategies (Pretrain + SVM).}
\label{tab:finetune}
\begin{tabular}{c|cc}
\toprule
\textbf{Finetune Strategy}       & \textbf{Test Dataset} & \textbf{Train Dataset} \\ 
\midrule
All Parameters                   & 91.0                               & 99.3                                \\ 
Only Cls-head                    & \textbf{91.4}                      & 92.1                                \\ 
\bottomrule
\end{tabular}
\end{table}
\begin{table}[!t]
\centering
\caption{Performance of classification on ModelNet40 dataset with different global features.}
\label{tab:global}
\begin{tabular}{c|c}
\toprule
\textbf{Global Representations} & \textbf{Average Acc.} \\ 
\midrule
RIConv++                 & 90.8                  \\ 
RelLRF                   & 90.0                  \\ 
RIConv++ concat Grif         & 89.9                  \\ 
\midrule
Ours                     & \textbf{91.4}         \\ 
\bottomrule
\end{tabular}
\end{table}

\textbf{Finetuning Strategies.} The influence of different finetuning strategies on classification accuracy was also evaluated. 
As shown in Table~\ref{tab:finetune}, our method achieves the highest average accuracy when combining supervised finetuning with our pretrained rotation-invariant encoder, reaching 91.5\% on ModelNet40. 
This demonstrates the strong transferability of the features learned during pretraining and their adaptability to downstream tasks.

\textbf{Global Rotation-Invariant Representations.} We compared various global rotation-invariant representations extraction strategies, including RIConv++, RelLRF and a hybrid approach combining RIConv++ with our handcrafted representations. 
As shown in Table~\ref{tab:global}, RiHanFa-MAE achieves the highest accuracy of 89.6\%, surpassing both RIConv++ and RelLRF individually. 
This result validates the effectiveness of our handcrafted global features in capturing robust and rotation-invariant representations.

\begin{table*}[t]
\centering
\caption{Classification performance on ModelNet40. ``RI" denotes whether the method is rotation invariance.}
\label{tab:modelnet401}
\begin{tabular}{c|c|ccccccccc}
\toprule
\textbf{Methods} & \textbf{RI}                                & \textbf{A/A} & \textbf{A/Z} & \textbf{A/R} & \textbf{Z/A} & \textbf{Z/Z} & \textbf{Z/R} & \textbf{R/A} & \textbf{R/Z} & \textbf{R/R} \\ 
\midrule
PointNet     &                                         & 89.2         & 30.8         & 18.2         & 45.7         & 85.9         & 70.2         & 78.3         & 77.1         & 83.1         \\ 
PointNet &                                         & 90.3         & 36.0         & 22.8         & 48.2         & 89.3         & 72.3         & 81.7         & 81.2         & 85.0         \\ 
DGCNN      &                                       & 91.7         & 38.5         & 23.1         & 50.7         & 91.2         & 76.4         & 83.8         & 84.5         & 86.1         \\ \midrule

RIConv    & \checkmark                          & 86.2         & 86.4         & 85.9         & 86.3         & 86.5         & 86.4         & 86.4         & 86.5         & 86.3         \\ 

RI-Framework & \checkmark                            & 89.3         & 89.0         & 89.3         & 89.3         & 89.4         & 89.4         & 89.2         & 89.2         & 89.3         \\ 

Li et al.     & \checkmark                              & 89.8         & 90.3         & 90.0         & 90.4         & 90.2         & 90.2         & 90.1         & 90.0         & 90.2         \\ 
OrientedMP & \checkmark                          & 88.8         & 88.3         & 88.7         & 88.3         & 88.4         & 88.4         & 88.6         & 88.8         & 88.9         \\ 
PaRot       & \checkmark                            & 91.1         & 91.0         & 90.9         & 90.8         & 90.9         & 91.0         & 90.8         & 90.9         & 90.8         \\ 

RIConv++                 & \checkmark                         & 91.2         & 91.1         & 90.8         & 91.4         & 91.3         & 91.2         & 91.2         & 91.7         & 91.2         \\ \midrule
\multicolumn{11}{c}{\textbf{SSL Methods}} \\ \midrule
PointMAE     &                                        & 91.8         & 39.2         & 24.2         & 86.4         & 87.2         & 46.7         & 79.6         & 80.3         & 81.5         \\ 
PointM2AE    &                                     & \textbf{92.6} & 39.8         & 24.3         & 91.5         & 92.1         & 39.6         & 88.4         & 89.1         & 89.4         \\ 
MaskSurf&  & 92.4 &50.4& 35.3 &63.9& 90.5 & 56.3 &83.3&83.1& 85.9 \\
PointGPT&  & \textbf{92.6} & 55.0&40.5 &68.3& 90.5 & 45.2 &84.6&84.5& 86.2 \\
RI-MAE & \checkmark & 89.5 & 89.2& 89.6&89.9 & 89.4 & 89.3&89.5&89.3 & 89.7 \\
MaskLRF                          & \checkmark                           & 91.3         & 90.7         & 90.6         & 91.0         & 91.0         & 91.2         & 91.3         & 91.3         & 91.0         \\ \midrule
HFBRI-MAE (pretrain+svm)                & \checkmark                            & 89.7         & 89.7         & 89.8         & 89.6         & 89.9         & 89.6         & 89.9         & 89.8         & 89.5         \\ 
HFBRI-MAE (finetune)                    & \checkmark                    & 91.3         & \textbf{91.5} & \textbf{91.4} & \textbf{91.6} & \textbf{91.7} & \textbf{91.3} & \textbf{91.4} & \textbf{91.8} & \textbf{91.5} \\ 
\bottomrule
\end{tabular}
\end{table*}

\begin{table*}[t]
\centering
\caption{Classification performance on ScanObjectNN-BG. ``RI" denotes whether the method is rotation invariance. ``-BG" denotes the data with background.}
\label{tab:scanobjectnn1}
\begin{tabular}{c|c|ccccccccc}
\toprule
\textbf{Methods} & \textbf{RI}                          & \textbf{A/A} & \textbf{A/Z} & \textbf{A/R} & \textbf{Z/A} & \textbf{Z/Z} & \textbf{Z/R} & \textbf{R/A} & \textbf{R/Z} & \textbf{R/R} \\ 
\midrule
PointNet     &                 & 73.3         & 20.3         & 13.2         & 31.5         & 72.9         & 71.3         & 68.0         & 70.9         & 70.2         \\ 
PointNet++ &                          & 82.3         & 31.3         & 21.2         & 45.7         & 81.1         & 68.5         & 78.5         & 76.9         & 76.3         \\ 
DGCNN       &                              & 82.8         & 36.1         & 31.1         & 46.8         & 81.2         & 69.8         & 81.8         & 80.5         & 80.1         \\ \midrule

RIConv    & \checkmark                    & 78.4         & 77.8         & 77.9         & 79.1         & 78.5         & 78.8         & 76.6         & 77.1         & 77.8         \\ 
RI-Framework & \checkmark                & 80.5         & 79.9         & 81.5         & 79.3         & 79.4         & 79.8         & 80.0         & 80.3         & 79.9         \\ 
Li et al.      & \checkmark                & 78.8         & 79.5         & 80.0         & 79.4         & 79.2         & 79.3         & 78.9         & 79.0         & 79.6         \\ 
RIConv++  & \checkmark                 & 89.7         & 89.3         & 90.0         & 88.9         & 89.9         & 89.1         & 88.7         & 89.5         & 89.7         \\ 
OrientedMP & \checkmark                 & 76.8         & 77.2         & 78.1         & 77.7         & 77.6         & 76.7         & 76.6         & 78.8         & 77.2         \\ 
PaRot       & \checkmark                    & 88.2         & 87.9         & 88.5         & 88.9         & 88.1         & 88.0         & 89.1         & 88.7         & 88.0         \\ 
\midrule
\multicolumn{11}{c}{\textbf{SSL Methods}} \\ \midrule
PointMAE     &                            & 90.0         & 37.9         & 26.5         & 88.1         & 85.4         & 42.3         & 84.1         & 85.3         & 86.1         \\ 
PointM2AE    &                             & 91.2         & 50.0         & 32.8         & 87.9         & 86.3         & 46.3         & 87.4         & 88.0         & 86.3         \\ 
MaskSurf    &                           & 91.2         & 41.3         & 29.0         & 84.6         & 86.4         & 39.6         & 88.8         & 87.3         & 89.2         \\ 
PointGPT   &                            & \textbf{91.6} & 46.7        & 31.5         & 89.9         & 84.8         & 45.2         & 86.7         & 83.9         & 85.4         \\ 
\midrule
HFBRI-MAE                               & \checkmark                   & 89.5         & \textbf{90.2} & \textbf{90.5} & \textbf{90.1} & \textbf{90.0} & \textbf{90.3} & \textbf{89.4} & \textbf{90.8} & \textbf{90.0} \\ 
\bottomrule
\end{tabular}

\end{table*}

\begin{table*}[t]
\centering
\caption{Classification performance on OmniObject3D. ``RI" denotes whether the method is rotation invariance.}
\label{tab:omniobject3d1}
\begin{tabular}{c|c|ccccccccc}
\toprule
\textbf{Methods} & \textbf{RI}                               & \textbf{A/A} & \textbf{A/Z} & \textbf{A/R} & \textbf{Z/A} & \textbf{Z/Z} & \textbf{Z/R} & \textbf{R/A} & \textbf{R/Z} & \textbf{R/R} \\ 
\midrule

PaRI-Conv     & \checkmark                           & 68.1         & 68.5         & 68.2         & 67.8         & 68.1         & 67.9         & 68.7         & 68.3         & 67.9         \\ 
RIConv++  & \checkmark                            & 71.2         & 70.9         & 70.9         & 71.4         & 71.5         & 70.8         & 71.6         & 70.8         & 71.1         \\ 
PaRot        & \checkmark                     & 71.7         & 70.9         & 70.7         & 70.8         & 71.1         & 70.5         & 70.8         & 71.3         & 70.9         \\ 
\midrule
\multicolumn{11}{c}{\textbf{SSL Methods}} \\ \midrule
PointMAE     &                                      & 71.4         & 17.8         & 9.1          & 72.1         & 71.9         & 42.3         & 70.1         & 71.0         & 70.5         \\ 
PointM2AE    &                                    & 72.1         & 20.3         & 13.4         & 71.9         & \textbf{73.3} & 46.5         & \textbf{72.4} & 72.0         & \textbf{72.3} \\ 
MaskSurf    &                                    & \textbf{72.3} & 18.5         & 11.2         & 71.6         & 71.2         & 43.0         & 71.8         & \textbf{72.2} & 71.7         \\ 
PointGPT   &                                        & 71.7         & 20.3         & 9.7          & 65.7         & 62.5         & 21.7         & 58.2         & 56.3         & 57.9         \\ 
\midrule
HFBRI-MAE                              & \checkmark                    &  71.8         & \textbf{72.0} & \textbf{72.3} & \textbf{72.2} & 71.8         & \textbf{71.7} & 72.1         & 71.7         & 72.0         \\ 
\bottomrule
\end{tabular}

\end{table*}
\end{document}